\documentclass[a4paper,twoside]{article}

\usepackage{epsfig}
\usepackage{subfigure}
\usepackage{calc}
\usepackage{amssymb}
\usepackage{amstext}
\usepackage{amsmath}
\usepackage{amsthm}
\usepackage{multicol}
\usepackage[dvipsnames]{xcolor}
\usepackage{pslatex}
\usepackage{apalike}
\usepackage{cite}
\usepackage{SCITEPRESS}     

\begin{document}

\title{$k$-fold Subsampling based Sequential Backward Feature Elimination}

\author{\authorname{Jeonghwan Park, Kang Li and Huiyu Zhou}
\affiliation{School of Electronics, Electrical Engineering and Computer Science, Queen's University Belfast, Belfast, United Kingdom}
\email{jpark04@qub.ac.uk, k.li@qub.ac.uk, h.zhou@ecit.qub.ac.uk}
}


\abstract{We present a new wrapper feature selection algorithm for human detection. This algorithm is a hybrid feature selection approach combining the benefits of filter and wrapper methods. It allows the selection of an optimal feature vector that well represents the shapes of the subjects in the images. In detail, the proposed feature selection algorithm adopts the $k$-fold subsampling and sequential backward elimination approach, while the standard linear support vector machine (SVM) is used as the classifier for human detection. We apply the proposed algorithm to the publicly accessible INRIA and ETH pedestrian full image datasets with the PASCAL VOC evaluation criteria. Compared to other state of the arts algorithms, our feature selection based approach can improve the detection speed of the SVM classifier by over 50\% with up to 2\% better detection accuracy. Our algorithm also outperforms the equivalent systems introduced in the deformable part model approach with around 9\% improvement in the detection accuracy. The source code is available at \textcolor{blue}{https://github.com/jpark04-qub/RSbSBE}}

\onecolumn \maketitle \normalsize \vfill


\section{\uppercase{Introduction}}
\label{sec:introduction}
\noindent A feature is an individual measurable property of a process being observed. Using a set of features, a machine learning algorithm can perform necessary classification \cite{Girish}. Compared to the situation back to the early years in the pattern recognition community, the space of features to be handled has been significantly expanded. High dimensionality of a feature vector is known to decrease the machine learning performance \cite{Guyon}, and directly affects applications such as human detection systems whose system performance relies heavily on both the classification speed and accuracy. A feature with no association with a class is regarded as a redundant or irrelevant feature. A redundant feature represents a feature which does not contribute much or at all to the classification task. An irrelevant feature can be defined as a feature which may only lead to decreased classification accuracy and speed. Blum (\textit{Blum}, 1997) defined the relevant feature $f$ as a feature which is useful to a machine learning algorithm $L$ with respect to a subset of features $\{S\}$: the accuracy of an hypothesized algorithm using the feature set $\{f \cup S\}$ is higher than that only using $\{S\}$.
In pattern recognition, the aim of feature selection is to select relevant features (an optimal subset), which can maximise the classification accuracy, from the full feature space. When a feature selection process is applied to pattern recognition, it can be seen as an embedded automated process which removes redundant or irrelevant features, and selects a meaningful subset of features. Feature selection offers many benefits in understanding data, reducing computational cost, reducing data dimensionality and improving the classifier's performance \cite{Girish}. In general, feature selection is distinguished from vector dimensionality reduction techniques such as Principal Component Analysis (PCA) \cite{Alpaydin}, as feature selection merely selects an optimal subset of features without altering the original information.
Feature selection requires a scoring mechanism to evaluate the relevancy of features to individual classes. The scoring mechanism is also named the feature selection criterion, which must be followed by an optimal subset selection procedure. Naively evaluating all the subsets of features ($2^N$) becomes an NP-hard problem (Amaldi and Kann, 1998) as the number of features grows, and this search becomes quickly computationally intractable. To overcome this computation problem, a wide range of search strategies have been introduced, including best-first, branch-and-bound, simulated annealing and genetic algorithms \cite{Kohavi}\cite{Guyon}, etc. In terms of feature scoring, feature selection methods have been broadly categorised into filter and wrapper methods \cite{Kohavi}. Filter methods allow one to rank features using a proxy measure such as the distance between features and a class, and select the most highly ranked features to be a candidate feature set. Wrapper methods score features using the power of a predictive model to select an optimal subset of features. \\
\indent In this paper, we propose a hybrid feature selection approach which combines the benefits of filter and wrapper methods, and discuss the performance improvement of the human detection system achieved using the proposed algorithm. We also demonstrate how optimal features selected by the proposed algorithm can be used to train an appearance model. 
The rest of the report is organised as follows. In section 2, feature selection methods and their core techniques are briefly reviewed. Section 3 introduces the proposed feature selection algorithm. The experiment results of the pedestrian detection system using the proposed algorithm is described in Section 4. Finally, Section 5 concludes the paper.

\begin{figure}[!t]
\centering
{\epsfig{file = 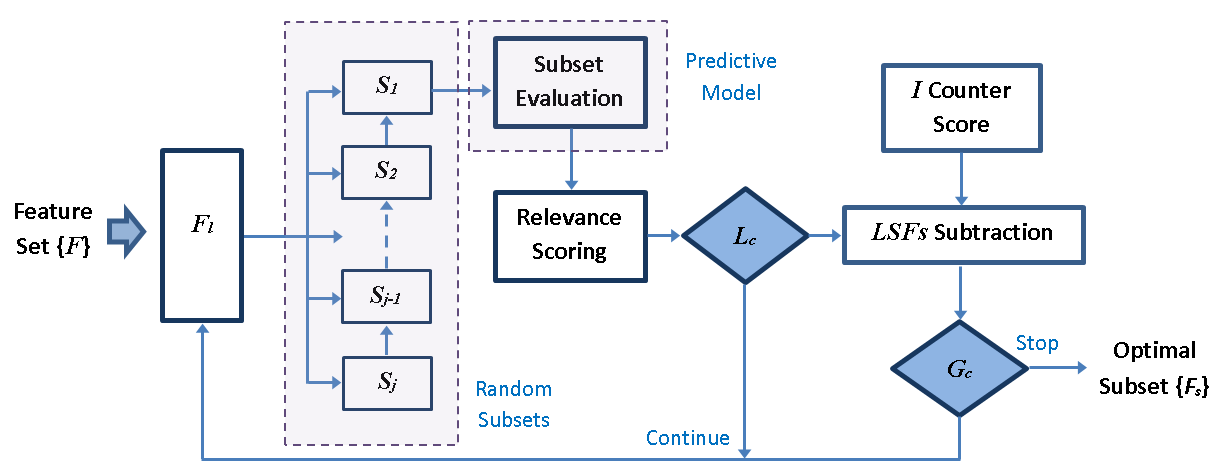, width = 7.5cm}}
\caption{Proposed Feature Selection System Overview.} 
\label{F:FS_Overview}
\end{figure}

\section{\uppercase{Related Work}}
\noindent One way for feature selection is simply evaluating features based on their information content, using measures like interclass distance, statistical dependence or information-theoretic measures  \cite{Estevez}. The evaluation is independently performed against different features, and the evaluation result called ``feature rank" is directly used to define the usefulness of each feature for classification. Entropy and Mutual information are popular ranking methods to evaluate the relevancy of features \cite{S_Foithong}\cite{H_Peng}\cite{Javed}\cite{Estevez}. \textit{Zhou et al.} \cite{Zhou_H} used the $R\acute{e}nyi$ entropy for feature relevance evaluation of overall inputs in their automatic scaling SVM. \textit{Battiti} proposed the MIFS algorithm \cite{MIFS}, which selects the most relevant \textit{k} feature elements from an initial set of \textit{n} feature dimensions, using a greedy selection method. Many MIFS variations have been introduced since then such as the mRMR \cite{H_Peng}, which used the first-order incremental search mechanism to select the most relevant feature element at a time. $\textit{Estevez et al.}$ \cite{Estevez} replaced the mutual information calculation in the MIFS by the minimum entropy of two features. \\
\indent Wrapper methods utilise classifier's performance to evaluate feature subsets. Wrapper methods have a significant advantage over filter methods as the classifier (learning machine) used for evaluation is considered as a black box. This flexible framework was proposed in \cite{Kohavi}. \textit{Gutlein et al.} \cite{Gutlein} proposed to shortlist \textit{k} ranked feature elements firstly, and then applied a wrapper sequential forward search over the features. \textit{Ruiz et al.} \cite{IWSS} proposed an incremental wrapper-based subset selection algorithm (IWSS), which identified the optimal feature area before the exhaustive search was applied.  \textit{Bermejo et al.} \cite{Bermejo} improved the IWSS by introducing a feature replacement approach. \textit{Foithong et al.} \cite{S_Foithong} used the CMI algorithm to guide the search of an optimal feature space, and applied the VPRMS as an intermediate stage before the wrapper method started. \textit{Pohjalainen et al.} \cite{J_Pohjalainen} proposed the RSFS which used the dummy feature relevance score as a selection criterion. \textit{Li and Peng} \cite{Kang_Li} introduced a fast model-based approach to select a set of significant inputs to a non-linear system. \textit{Heng et al.} \cite{C.K.Heng} addressed the overfitting problem of the wrapper methods by proposing a shrink boost method. \textit{Yang et al.} \cite{Y.Yang} proposed a wrapper method with the LRGA algorithm to learn a Laplacian matrix for the subset evaluation. 

\begin{figure*}[!t]
\centering
\includegraphics[width=5.5in]{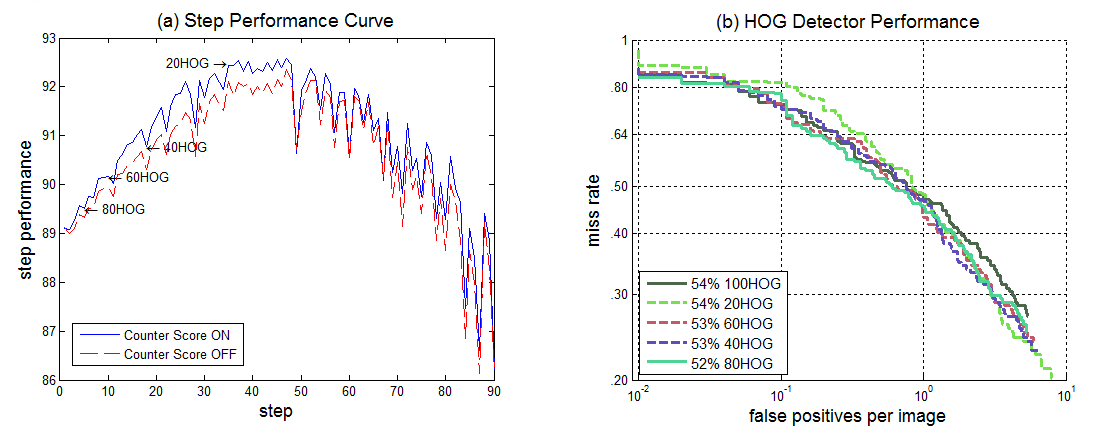}
\caption{Performance illustration: (a) Step performance curve of the proposed feature selection algorithm. (b) Human Detection performance using different HOG detectors trained with different feature subsets. The evaluation was carried out on the first 100 INRIA Full Images. The computed miss rates between 0.01 and 1 false positives per-image (FPPI) are shown in the legend. HOG Detectors trained with optimal feature subsets perform better than or similar to 100HOG showing a small overfitting problem.}
\label{F:HOG_Detectors}
\end{figure*}

\section{\uppercase{Proposed Algorithm}}
\noindent In this section, a novel feature selection algorithm is presented. The proposed feature selection algorithm is a wrapper method which adopts the \textit{k}-fold subsampling method in the validation step. The search strategy is an exhaustive search, namely the sequential backward elimination (SBE, \textit{Marill and Green} 1963). The proposed algorithm is a classifier dependent feature selection method, and the linear SVM is used as the classifier to evaluate the selected feature subsets. 

\subsection{Frame Work}
\noindent First, the entire feature set $\{F\}$ is randomly divided into $j$ small subsets where an evaluation process is performed, $S_{n} \in F, \; for \; n = 1, 2, .., j$. Square root calculation on the size of the full feature set $\{F\}$ is used to determine the size of a subset. When the local stopping criterion $L_c$ is satisfied, another relevance score $I$ contributes to the computation of feature relevance scores, and considerably a large number of irrelevant/redundant feature elements (from the least significant one, $LSFs$) are subtracted. The process continues on until the global stopping criterion is satisfied. The whole process is sketched in Fig.\ref{F:FS_Overview}. \\

\textbf{Random Subsets :} 
When the remaining feature set $\{F_l\}$ is reset, a temporary ID is given to individual vector elements for backtracking which gives them an equal opportunity in the evaluation. The temporary ID is only valid within one step. A step refers to the point where the local stopping criterion $L_c$ is satisfied, $LSFs$ are subtracted and all the iteration parameters are re-initialised. An iteration refers to that the evaluation has completed over all the subsets, and each vector element has been evaluated once. At the beginning of each iteration within one step, the feature vector elements of the remaining feature set $\{F_l\}$ is randomly re-arranged and divided into subsets, $n \times \{S\}$. The size of a subset is chosen as $\sqrt{N}$ where $N$ is the number of the remaining features.  \\

\textbf{Relevance Score :}
The algorithm uses two scores, step performance score $P$ and feature relevance score $R$. During each iteration in the exhaustive search stage, the relevance score of individual features is updated according to the prediction power score over the subset where the feature participates. In this paper, the Unweighted Average Recall (UAR) is used to calculate the prediction power score over the subsets:
\begin{equation} \label{eq:UAR}
\begin{aligned}
P\{S_n\} = \frac{1}{N}\sum\limits_{i=1}^N \Big(\frac{C_{iP}}{C_{iP} + C_{iF}} \Big)
\end{aligned}
\end{equation}
where $S_n$ is $n$th subset, $i$ is the class index, $C_{iP}$ is the class true positive (correct prediction) and $C_{iF}$ is the class false negative (wrong prediction). \\
\indent The individual feature relevance score is then updated as \cite{J_Pohjalainen}:
\begin{equation} \label{eq:relevance_score}
\begin{aligned}
R_f = R_f + P\{S_n\} - E, \quad f \in S_n,
\end{aligned}
\end{equation}
where $E$ is the UAR of the cumulated $C_P$ and $C_F$ over a step.
As the search continues, the feature score $R_f$ represents how much the corresponding feature has contributed to the prediction. \\

\textbf{Stopping Criterion :} 
The local stopping criterion $L_c$ is calculated using the standard deviation of the step's performance with a specific predictor. $L_c$ is defined as:
\begin{equation} \label{eq:local_criterion}
\begin{aligned}
L_c = \sqrt{\frac{1}{N}\sum\limits_{i=1}^{N}\Big(P_i - \frac{1}{N}\sum\limits_{j=1}^{N} P_j\Big)^2}
\end{aligned}
\end{equation}
where $P_i$ is the step performance score. $L_c$ is then compared with a supervised parameter (0.6 in this study) to decide the evaluation fairness over all the features. 
The global stopping criterion $G_c$ is a supervised parameter, and the number of features to be finally selected is used here. \\

\begin{figure*}[!t]
\centering
\includegraphics[width=6in]{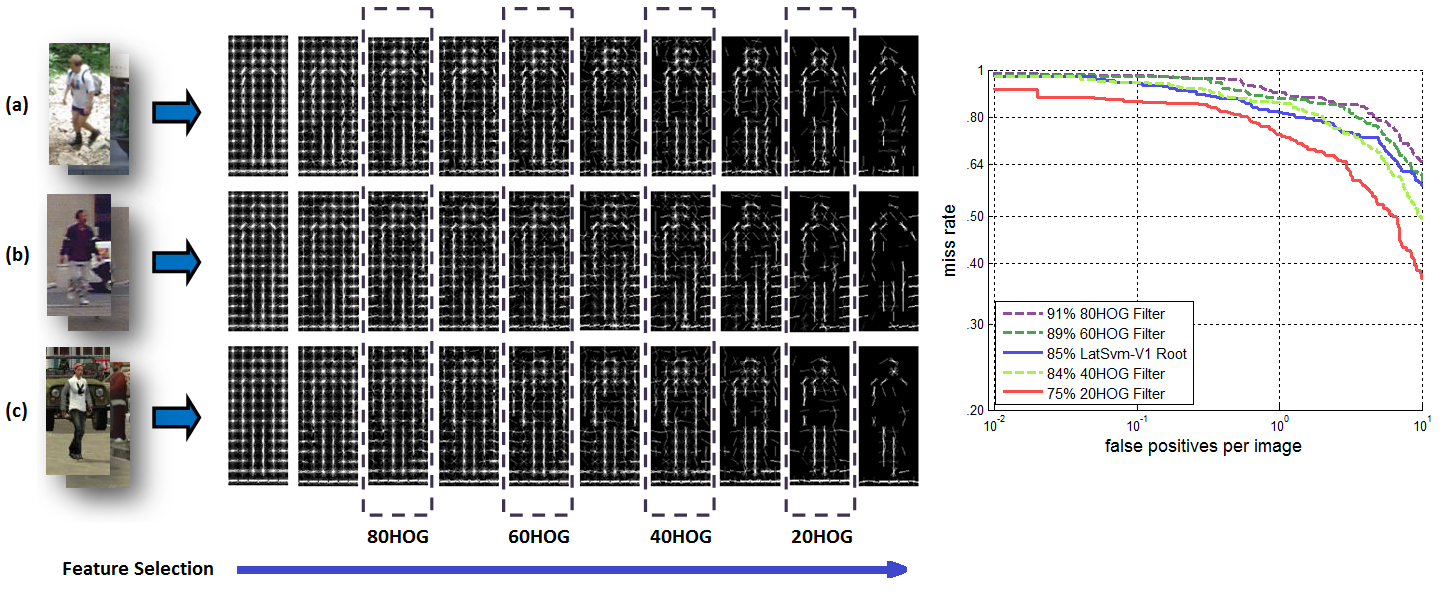}
\caption{Appearance models trained using the proposed feature selection algorithm. (a) INRIA Dataset, (b) MIT Dataset and (c) CVC04 Dataset \cite{CVC04}. The detection performances of the INRIA appearance models are evaluated on the first 100 INRIA full images. The appearance model from 20HOG shows 10\% better performance than the root filter of LatSVM-V1 \cite{LatSvm-V1}.} 
\label{F:Appearance_Models}
\end{figure*}
\textbf{Counter Score :}
Even though an algorithm has performed a large number of iterations and the local stopping criterion $L_c$ has been satisfied, it can not always be guaranteed that the chosen $LSFs$ are truly irrelevant/redundant features, especially in a large feature space. To overcome this problem, the proposed algorithm uses information ranks, named \textit{counter score}, in the $LSFs$ selection to update each feature's relevance score. In this paper, the mutual information is chosen to compute the counter score.
Mutual information was originally introduced in information theory. It is used for empirical estimates between individual features and classes \cite{Guyon}. Mutual information is derived from entropy. Entropy $H$ is an uncertainty measure of event occurrence. Entropy of the discrete random variable $X$ is described as $H(X) = -\sum\nolimits_{x\in X} p(x) \log{p(x)}$, where $p(x)$ denotes the probability density of an event $x\in X$. The entropy of variable $X$ can be conditioned on variable $Y$ as $H(X|Y)$.
If variable $Y$ does not introduce any information which influences the uncertainty of $X$, in other word, $X$ and $Y$ are statistically independent, the conditional entropy is maximised \cite{Jorge_R}. 
From this description, mutual information $IG(X;Y)$ can be derived as $H(X) - H(X|Y)$. Mutual information represents the amount of information mutually shared between variable $X$ and $Y$. This definition is useful within the context of feature selection because it gives a way to quantify the relevance of a feature with respect to the class \cite{Jorge_R}. Therefore, using mutual information in the wrapper approach benefits both the optimal feature space search and the selection performance enhancement. The proposed algorithm uses the mutual information to compute the counter score of each feature. The counter score $I_f$ and its contribution to the feature relevance score are calculated as follows: 
\begin{equation} \label{eq:MI_contribution}
R_f = R_f + \alpha I_f, \\
\quad I_f = IG_f/IG_{Max}
\end{equation}
where $\alpha = (R_{Max} \times FN_{Remain})/FN_{Full}$ is the counter score contribution rate, $IG_f$ is the mutual information of feature elements. The rate is dynamically decremented as more steps are processed $\frac{FN_{Remain}}{FN_{Full}}$, which means the counter score contributes more in the large $LSFs$ subtraction. Fig.\ref{F:HOG_Detectors} (a) shows that the counter score improves the performance of the proposed wrapper feature selection algorithm in terms of local prediction accuracy.   \\

\textbf{Feature Subtraction :}
The number of features to be removed at each step is chosen as $(5/N)\times 100$, where $N$ is the number of the remaining features. In comparison to the original SBE algorithm, which removes only the least significant feature at a time, it is reasonable to remove more than one feature at a time, as only a small portion of features are highly relevant in many applications. In a human detection system, the most relevant features can be viewed as the features which are centred on the human contour. This can be visually demonstrated in Fig.\ref{F:Appearance_Models}.
When a step is completed ($L_c$ has been satisfied), the algorithm subtracts a group of the least significant features, $m \times LSFs$, which have the lowest relevance score $R_f$.   \\

\begin{figure*}[!t]
\centering
\includegraphics[width=6.2in]{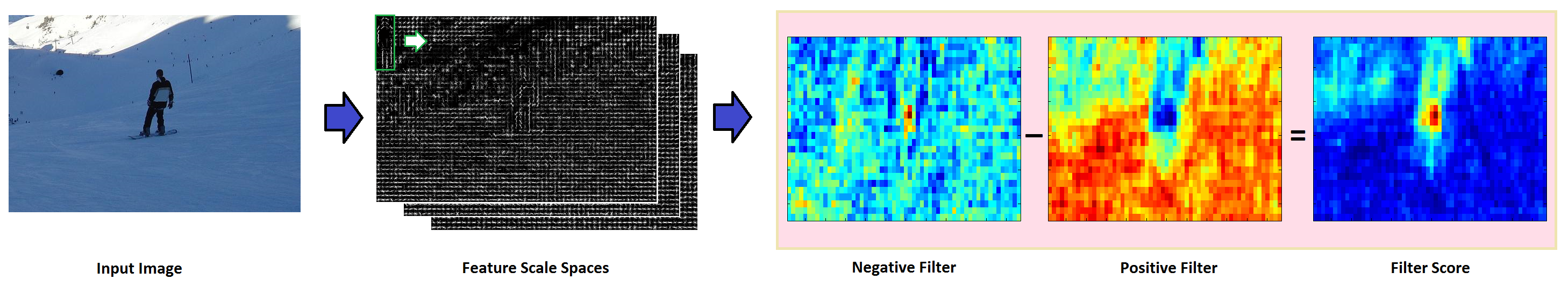}
\caption{Filter Score Computation: The positive filter score is subtracted from the negative filter score.}
\label{F:Filter_Score}
\end{figure*}   

\subsection{Appearance Model}
\noindent In a human detection system, the optimal feature elements tend to represent the human contour as illustrated in Fig.\ref{F:Appearance_Models}. This was also pointed out in \cite{N_Dalal} as the most important cells are the ones that typically contain major human contours. In other words, the optimal feature elements are useful to build an appearance model. To evaluate the discriminative power of the selected feature elements, a simple appearance model is created. The model consists of a positive filter and a negative filter. The positive filter is formed with the average HOG of the selected feature elements from the positive examples. The negative filter is generated with the negative examples in the same manner. 
The HOG scheme allows us to divide the appearance of an object into geometrical units called cells. A cell is then represented with $n$ angles and their weighted magnitudes. The HOG feature of an example is an one-dimensional array of magnitudes whose indexes indicate the angle and the location of the cell. The proposed feature selection algorithm outputs the array indexes of an optimal feature subset. Each element $f_i$ of a filter is computed as follows:   
\begin{equation} \label{Eq:Filter_Train}
\begin{aligned}
f_i = \frac{1}{n}\sum\limits_{j=1}^{n}\beta_{ji}
\end{aligned}
\end{equation}
where $f_i$ is the array element of the proposed filter, $\beta$ is the vector of an example of dataset.   
The score of region $S_r$ is the regional similarity and computed with the euclidean distances between the ROI and the filters as shown in the following Equation:
\begin{equation} \label{Eq:Filter_Score}
\begin{aligned}
S_r = d(N_s,O_s) - d(P_s,O_s)
\end{aligned}
\end{equation}
where $N_s$ and $P_s$ are the optimal sub-vectors of the negative and positive filters, $O_s$ is the sub-vector of ROI feature vector, and $d(x,y)$ is the euclidean distance. 

\begin{figure*}[!t]
\centering
\includegraphics[width=6in]{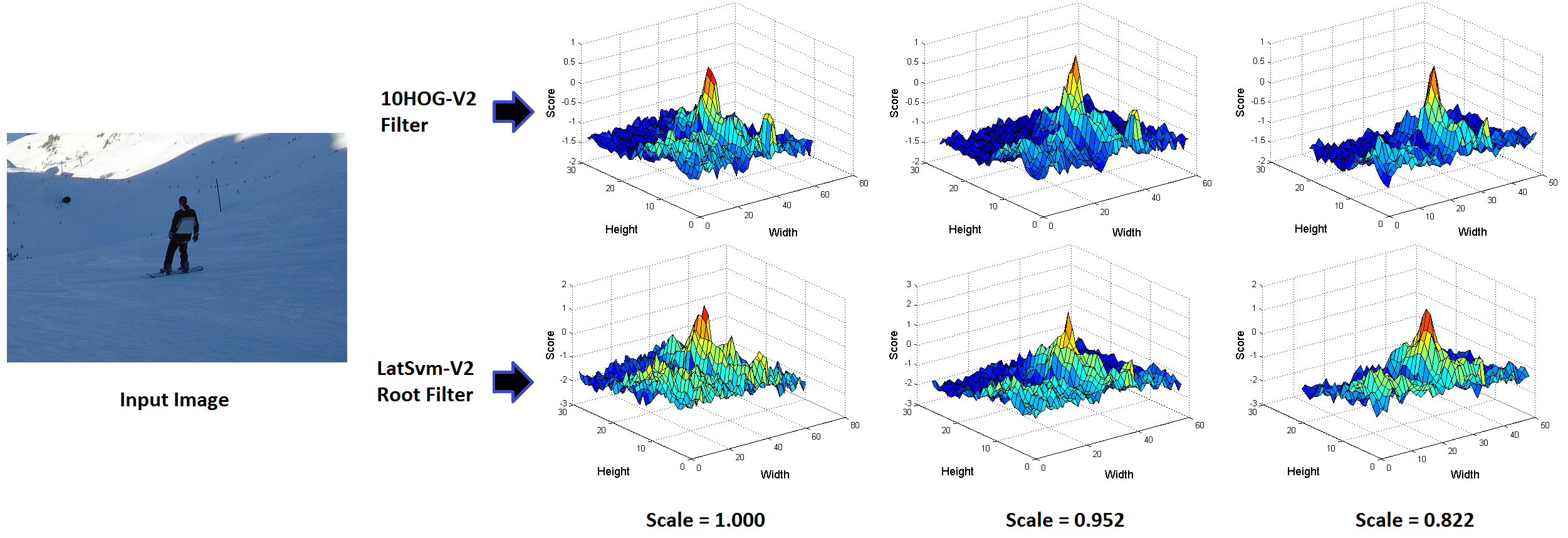}
\caption{Localisation accuracy: The human localisation accuracy of the 10HOG-V2 Filter is compared to that of the root filter of the LatSvm-V2 Model \cite{LatSvm-V2}. Top - Score maps of the 10HOG-V2 filter at three scales. Bottom - Score maps from the LatSvm-V2 root filter at the identical scale. The score map from the 10HOG-V2 filter shows less noise than the one of the LatSvm-V2 root filter.}
\label{F:Localisation_Compare}
\end{figure*}   

\section{Experimental Work}
\noindent $\textit{Doll\'ar}$ \cite{Dollar2012PAMI} \cite{Dollar2009CVPR} provided 288 INRIA pedestrian full images for the benchmarking purpose. The majority of tests in this paper are carried out against 288 full images. However, some of the tests are conducted on the first 100 images, which show similar results as for the case of 288 full images. To evaluate the generalisation of the proposed approach, the test is also performed on the ETH Bahnhof sequence \cite{ETH}, which contains 999 street scenes. The performance of the human detection systems in terms of the detection accuracy are evaluated using the PASCAL VOC criteria \cite{PASCAL07}. All the algorithms and systems in the experiments are realised using Matlab 2014 with the relevant mex files. The test computer is of 2.49GHz Intel i5-4300U CPU running with Window 8. 

\subsection{Feature Vectors}
\textbf{Feature Vector:}
\noindent The feature used in the experiments is the Histograms of Oriented Gradients (HOG) \cite{N_Dalal}. 
First of all, a linear SVM classifier is trained using the MIT pedestrian dataset. The MIT dataset consists of up-straight person images which have less dynamic poses. The positive examples shortlisted from other datasets using this classifier tends to include rather static pose examples. The dataset consisting of static pose examples is used to build an appearance model. Secondly, a subset of the INRIA dataset is selected by the classifier. The INRIA dataset offers cropped 2416 positive examples, and also allows to generate 12180 negative examples in the random selection manner for the training purpose. The classifier selects 1370 positive and 1579 negative examples from the training dataset. The negative examples include 79 false positive examples called "hard negative example". 
Thirdly, the feature extraction algorithm introduced in \cite{LatSvm-V1} is used to compute HOG descriptors for the experiments with LatSvm-V1 (Feature Vector A). The algorithm in \cite{LatSvm-V2} is also used for the tests with LatSvm-V2 (Feature Vector B). \\

\textbf{Feature Selection:}
The proposed feature selection algorithm is applied to the extracted feature vectors shown above.
From Feature Vector A, the algorithm selects four optimal feature subsets which have 80\%, 60\%, 40\% and 20\% elements of the full feature vector. The detection system trained with these feature subsets are referred to as 80HOG, 60HOG, 40HOG, and 20HOG respectively. 100HOG represents the system trained with the full feature vector.
The algorithm also selects 20\%, 15\% and 10\% elements from Feature Vector B. They are referred to as 20HOG-V2, 15HOG-V2 and 10HOG-V2 respectively. 

\subsection{Full Image Results}
\textbf{Feature Vector A:}
To evaluate the feature selection performance, a simple human detection system using the HOG and the linear SVM \cite{N_Dalal} is created.  
Fig.\ref{F:HOG_Detectors} (b) shows the detection accuracy of the systems trained with the selected feature subsets from Feature Vector A. The 40HOG, 60HOG and 80HOG slightly improve the accuracy up to 2\% compared to the system trained with the full feature set, 100HOG. The detection accuracy is improved until the 40HOG is applied, which uses less than half of the full feature vector dimension. The 20HOG shows no improvement in the detection performance even though 20\% feature vector has the best score in the local classification score curve as shown in Fig.\ref{F:HOG_Detectors} (a). The results of the evaluation reveal that as the feature selection progresses the proposed algorithm gradually introduces the overfitting problem. The window sliding speed is significantly improved. The 100HOG takes 478.634s to scan $1060 \times 605$ pixels image. Compared to this, the 20HOG takes only one tenth of the search time required by the original system completing the same search within 45.895s.\\ 
\begin{figure*}[t]
\centering
\includegraphics[width=6in]{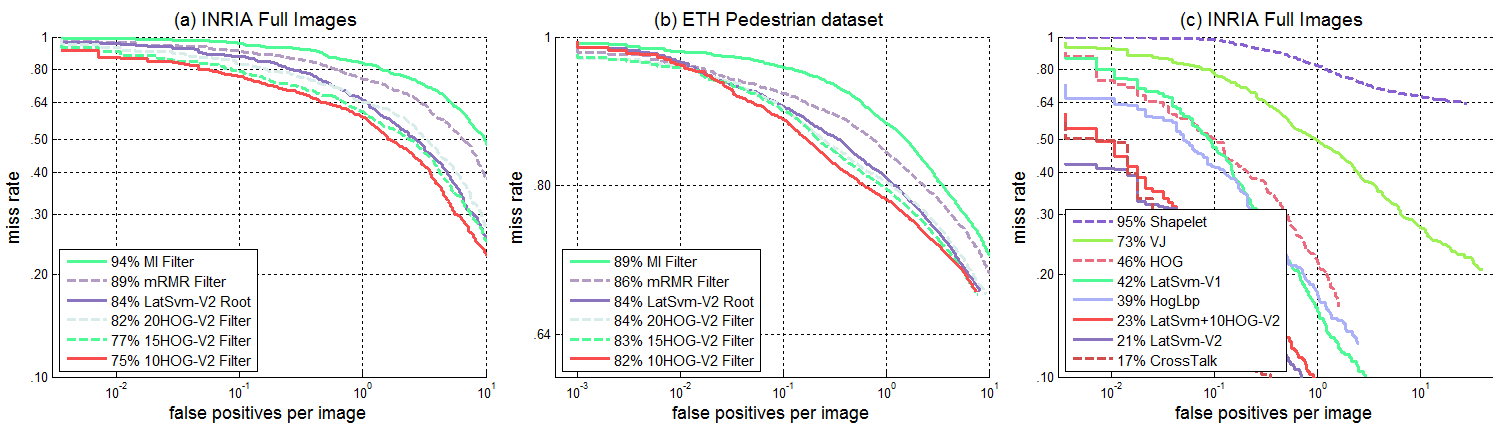}
\caption{Performance comparison: (a) Filter detection performance comparison on the 288 INRIA Full Images. (b) Filter detection performance on the ETH Bahnhof sequence, (c) Comparison to the state-of-the-arts on the 288 INRIA Full Images.}
\label{F:DPM-V2_Result}
\end{figure*}    
\textbf{Feature Vector B:}
The proposed appearance model is trained with the 20HOG-V2, 15HOG-V2 and 10HOG-V2. Fig.\ref{F:DPM-V2_Result}(a) shows the evaluation results of the appearance models. The evaluation is carried out on the 288 INRIA Full Images, and compared to the LatSvm-V2 Root filter \cite{LatSvm-V2} and the appearance models trained with the other feature selection algorithms.
Unlike the classifier dependent systems, the appearance model shows better performance as the feature vector is better optimised, and no overfitting problem is observed with the appearance models. The 10HOG-V2 model outperforms the LatSvm-V2 Root filter scoring with a 9\% better detection rate. 
To test the generalisation performance of the proposed approach, filters trained with the INRIA dataset are directly applied in the experiments on the ETH Bahnhof sequence \cite{ETH}, which consists of 999 street scenes. The 10HOG-V2 outperforms all the other filters achieving 2\% better performance compared to the LatSvm-V2 root filter as shown in Fig.\ref{F:DPM-V2_Result} (b).  
The root filter of the LatSvm approach \cite{LatSvm-V1}\cite{LatSvm-V2} is equivalent to the model from \textit{Dalal}'s original HOG approach \cite{N_Dalal}. Therefore, the proposed appearance model can simply replace the LatSvm root filter. Fig.\ref{F:DPM-V2_Result} (c) illustrates the detection performance evaluation of the Deformable Part Model, LatSvm+10HOG-V2, which has the proposed appearance model as a root filter. The evaluation on the 288 INRIA Full Images shows that the improved performance of the 10HOG-V2 filter is slightly worse than the LatSvm+10HOG-V2. 
The performance decrease can be explained in two-folds: Firstly, the part filters of the Deformable Part Model contributes more than the root filter does. On the first 100 INRIA Full Images, the part filters scores a 44\% detection rate, while the root filter achieves 91\%. Secondly, there are many supervised parameters involved in the Deformable Part Model, and the supervised parameters appear to affect the performance of the LatSvm+10HOG-V2. Therefore, the further optimisation is required to make the proposed filter fit with the Deformable Part Model.
\section{Conclusion}    
\noindent We have presented a feature selection algorithm which can generate the optimal feature subset. We have demonstrated the chosen feature subset can be used to improve the human detection system, which relies on the classifier performance, in both speed and accuracy. It has also been shown that the optimal features represent the object shape. Base on this observation, we have demonstrated that the optimal feature vector can be directly used to form the appearance models. This approach does not require highly accurate annotation data of objects to generate models. Therefore, it can be easily applied to a wide range of datasets. 

\bibliographystyle{apalike}
{\small
\bibliography{ICPRAM-Ref}}

\vfill
\end{document}